\documentclass[
]{ceurart}

\usepackage{listings}
\usepackage{caption} 
\usepackage{subcaption} 
\usepackage{graphicx} 

\begin{document}

\copyrightyear{2023}
\copyrightclause{Copyright for this paper by its authors. 
  Use permitted under Creative Commons License Attribution 4.0
  International (CC BY 4.0).}

\conference{HHAI-WS 2023: Workshops at the Second International Conference on Hybrid Human-Artificial Intelligence (HHAI), June 26—27, 2023, Munich, Germany.}

\title{Leveraging Diversity in Online Interactions}

\author[]{Nardine Osman}[%
orcid=0000-0002-2766-3475,
email=nardine@iiia.csic.es,
]
\address[]{Artificial Intelligence Research Institute (IIIA-CSIC), Barcelona}

\author[]{Bruno Rosell i Gui}[%
orcid=0000-0001-9960-8734,
email=rosell@iiia.csic.es,
]

\author[]{Carles Sierra}[%
orcid=0000-0003-0839-6233,
email=sierra@iiia.csic.es,
]


\begin{abstract}
  This paper addresses the issue of connecting people online to help them find support with their day to day problems. We make use of declarative norms for mediating online interactions, and we specifically focus on the issue of leveraging diversity when connecting people. We run pilots at different university sites, and the results show relative success in the diversity of the selected profiles, backed by high user satisfaction.
\end{abstract}


\maketitle

\section{Introduction}
Despite the growing use of generative AI in various domains~\cite{DWIVEDI2023102642}, connecting people to help them solve their problems and enrich their interactions will always be imperative. \cite{stackOverflow23} shows that Stack Overflow's traffic went down 14\% in March 2023, due to people moving to ChatGPT for finding answers. However, this is not sustainable, as ChatGPT finds `good' answers only because it trains on `good' data, such as that provided by Stack Overflow. We argue that data flow from us humans will always be key. Similarly, finding support in fellow humans will always be indispensable. 

The WeNet project attempts to address the issue of connecting people to help them find support with their day to day problems. While there are many platforms that do so today, leveraging AI to find suitable people to connect to is still lagging.  In this paper we focus on the issue of mediating humans' online interactions
to ensure online behaviour unfolds according to expectations. As we focus on humans and their needs, our work has been motivated by: 1) empowering users, by giving them control over some of the technology's functionalities, yet ensuring crucial basic requirements are not broken; 2) connecting people, by developing profile matching mechanisms that leverage profile diversity; and 3) aligning users' understanding of the norms to ensure community members have a clear understanding of the norms mediating their interactions. We thus contribute to the hybrid machine-artificial intelligence field through diversity-enriching interaction-mediating technologies. 

The following three sections (Sections \ref{sec:1}--\ref{sec:3}) describe our approach for addressing the above three issues. Section~\ref{sec:pilot} provides some initial results from our pilots. Finally, we conclude with Section~\ref{sec:conc} that analyses the results of the pilots and suggests future work.

\section{A Declarative Approach for Modelling Interactions}\label{sec:1}
The proposed WeNet interaction model~\cite{OsmanSCSG20,OsmanCSSG21} tackles the issue of empowering users by considering both, user requirements and system (or organisational) requirements. This gives users control over some functionalities (such as specifying what profiles are deemed relevant), while ensuring crucial basic requirements are not violated (such as ethical requirements and avoiding biases). These requirements that are mediating interactions are specified through declarative norms~\cite{normsG,norms2,2012-D'Inverno}.  

Following a declarative approach for norm representation allows for dynamic norms that can change at runtime. This has proven to be useful in our pilots as we could continue to tweak online behaviour (through norms) even after the app's deployment into production. For our plans for future work (Section~\ref{sec:conc}), we plan to learn and adapt the norms at runtime. This highlights further the need for a declarative approach. 

The language used to specify norms is based on a simple \textit{if then} statement structure. The conditions are selected from the developed application's variables, allowing the norms to check issues such as if a question has been created, answered, etc., or whether certain profile attributes satisfy certain requirements. Similarly, the consequences are also based on the the application's acceptable actions, such as sending messages, updating a profile, etc. 

\section{Leveraging Diversity when Selecting Profiles}\label{sec:2}
Connecting people is another of the main objectives of WeNet. We especially focus on leveraging diversity when selecting the profiles to tackle a given issue~\cite{abs-2005-09331,GeorgaraRS21,GeorgaraRSMKAP22}. However, we ask the person making a question, the questioner, to decide which diversity dimensions are relevant for their question. We believe that different diversity dimensions might be suited for different contexts. 

We choose to focus on four diversity dimensions: selecting profiles based on their domain interests, beliefs and values, social closeness, and physical closeness. Notice that we focus on what we refer to as deep features to avoid potential biases from arising when allowing the questioner to filter profiles based on shallow features like gender or nationality. Nevertheless, the WeNet project developed a diversification mechanism based on shallow features, to ensure that selected profiles are sufficiently diverse with respect to a given shallow feature, like gender.

How the above diversity dimensions are used in the different pilots to leverage diversity is is domain dependent, and it is the norms that dictate how those are used.

\section{Norm Alignment through User Feedback}\label{sec:3}
Ensuring community members have a clear and aligned understanding of the norms that mediate their interactions is another issue that we have focused on. Say there is a norm that prohibits hate speech. We argue that what is considered hate speech might change from one community to another, or even change over time. As such, we have developed learning mechanisms~\cite{thiago1, thiago2, thiago3} that allow the system to understand the meaning of norms (say what constitutes hate speech) from people's interaction and their feedback. It can then use what it learned to guide users' actions by highlighting when they violate certain norms and why. 


\section{Pilots}\label{sec:pilot}
We have developed a number of pilots across different sites: London School of Economics (LSE), Aalborg University (AAU), and the National University of Mongolia (NUM). The objective was for students to use our developed application to find other students that can help answer their questions. 

The diversity dimensions were defined differently for each of the pilot sites, and how the norms leveraged all diversity dimensions also varied from one pilot site to another, based on each site's requirements. For example, LSE gave different weights to the different dimensions while AAU and NUM did not, although they did specify that some diversity dimensions were primary while others were secondary, where requirements on primary dimensions should be satisfied first. Furthermore, social closeness was interpreted and implemented differently at each site. 

\begin{figure}[b]
\centering
    \begin{subfigure}[b]{0.45\textwidth}
        \centering
        \includegraphics[width=\textwidth]{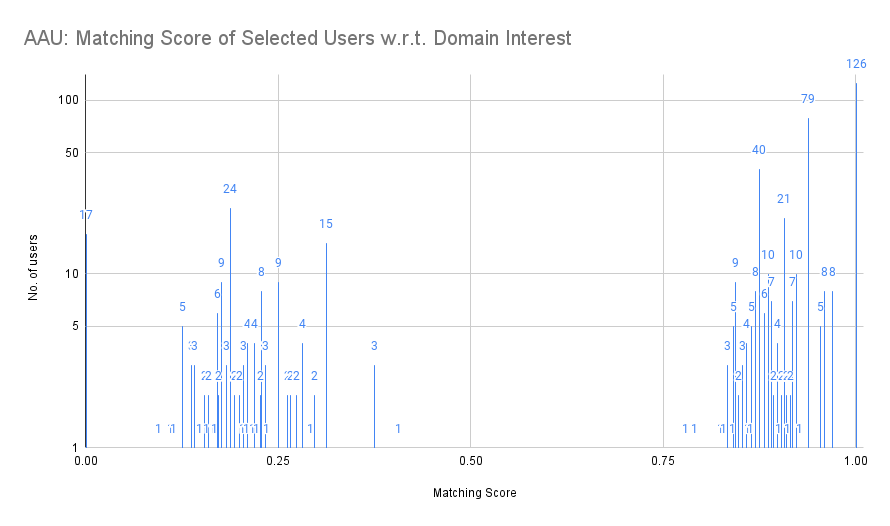}
        \caption{Satisfaction of diversity requirements w.r.t. domain interests}
        \label{fig:M46AAU_di}
    \end{subfigure}
    \hfill
    \begin{subfigure}[b]{0.45\textwidth}
        \centering
        \includegraphics[width=\textwidth]{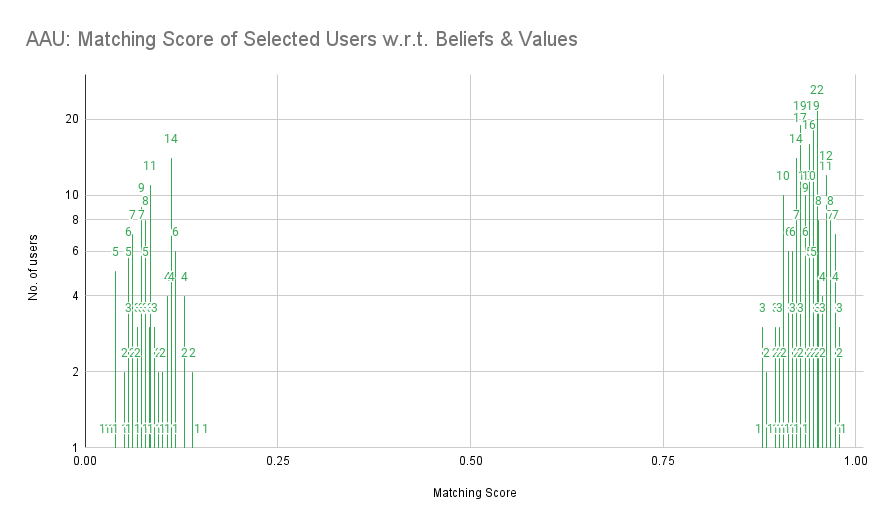}
        \caption{Satisfaction of diversity requirements w.r.t. beliefs and values}
        \label{fig:M46AAU_bf}
    \end{subfigure}

    \begin{subfigure}[b]{0.45\textwidth}
        \centering
        \includegraphics[width=\textwidth]{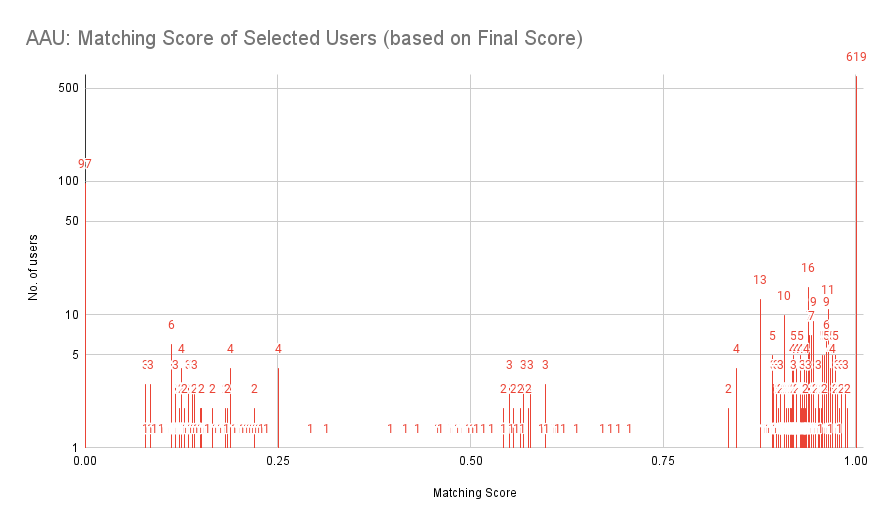}
        \caption{Satisfaction of diversity requirements in general}
        \label{fig:M46AAU_ams}
    \end{subfigure}
\caption{The selected profiles at the AAU pilot site and their satisfaction of diversity requirements}
\label{fig:resultsAAU}
\end{figure}

Figure~\ref{fig:resultsAAU} presents some of the results of the AAU pilot. The results of the other pilots are aligned with those of AAU. The $y$-axis presents the number of selected profiles with the corresponding matching score ($x$-axis), or the score that describes the degree of a profile of satisfying the diversity requirements. The graphs follow a logarithmic scale. 

The results highlight the capability of diversifying selected profiles using declarative norms. Different iterations of the pilots illustrated that the larger the communities, the easier it was to find profiles that fit the diversity requirements. We are confident that even larger community sizes would result in even better profile matches. (Community size in the last iteration whose results are presented here were at 105 users at LSE, 51 users at AAU, and 115 users at NUM.) Finally, the exit survey highlighted the users' satisfaction with the diversity dimensions and appreciated their value opportunity.

\section{Conclusions and Future Work}\label{sec:conc}

The pilots, first and foremost, showcased the power of a declarative approach to norm specification. Even when the deadline has passed for developing the app and moving it to production, we continued to be able to adapt some functionalities through declarative norms that can change at runtime.

The norms opened the door for users to specify their diversity requirements. For a given question to be put forward in a community, the questioner had the option to choose whether they were interest in looking for similar/diverse profiles with respect to domain interests and beliefs and values, as well as close/distance profiles with respect to social and physical distance. The norms also implemented some fixed requirements, such as sending a question to five users only, and diversifying the final list of potential responders with respect to gender. The results of these pilots showed that the larger the communities, the easier it is to leverage diversity.  

Explainability has emerged as a key requirement for users to understand how the norms and the diversity requirements function. The last pilot iteration demonstrated the success of explainability, though it also highlighted the challenges of explaining complex algorithms to users. A trade-off needs to be made when it comes to the details versus the clarity of the explanations.

The interest in leveraging diversity was limited, despite the fact that the exit survey has illustrated that users did in fact find value in those diversity requirements. As such, we strongly believe that diversity requirements are very much context dependent. When applicable, their value is appreciated. Though further research is needed to analyse the applicability per context.

Last, but not least, the pilots have highlighted the difficulty in predicting which norm interpretation is applicable for different contexts. For example, 
when should social closeness depend on past relations, and when should it depend on coming from similar academic circles? How many users should receive a question to guarantee good enough answers are collected, yet users are not bombarded with numerous questions? We believe that feedback is key in answering these questions. Furthermore, addressing those questions may be automated. 

We argue that our work on developing mechanisms that allows the system to understand the meaning of norms (say what hate speech means) from people's interaction and their feedback can provide the basis for the next steps in our research, which we believe should focus on using such mechanisms to learn the most suitable norm for a given community. For example, analysing interaction data 
and user feedback 
can help learn the best number of people to forward a question to in a given community. And this number can be adapted to different communities. One may also learn which meaning of social closeness do users prefer and under what context. This can help to automatically adapt the interpretation and specification of norms for different contexts.

\begin{acknowledgments}
  This research is supported by the EU funded VALAWAI (\#~101070930) and WeNet (\#~823783) projects, and the Spanish funded VAE (\#~TED2021-131295B-C31) and Rhymas (\#~PID2020-113594RB-100) projects.  
\end{acknowledgments}

\bibliography{bibliography}

\begin{thebibliography}{13}
\expandafter\ifx\csname natexlab\endcsname\relax\def\natexlab#1{#1}\fi
\providecommand{\url}[1]{\texttt{#1}}
\providecommand{\href}[2]{#2}
\providecommand{\path}[1]{#1}
\providecommand{\DOIprefix}{doi:}
\providecommand{\ArXivprefix}{arXiv:}
\providecommand{\URLprefix}{URL: }
\providecommand{\Pubmedprefix}{pmid:}
\providecommand{\doi}[1]{\href{http://dx.doi.org/#1}{\path{#1}}}
\providecommand{\Pubmed}[1]{\href{pmid:#1}{\path{#1}}}
\providecommand{\bibinfo}[2]{#2}
\ifx\xfnm\relax \def\xfnm[#1]{\unskip,\space#1}\fi
\bibitem[{Dwivedi et~al.(2023)Dwivedi, Kshetri, Hughes, Slade, Jeyaraj, Kar,
  Baabdullah, Koohang, Raghavan, Ahuja, Albanna, Albashrawi, Al-Busaidi,
  Balakrishnan, Barlette, Basu, Bose, Brooks, Buhalis, Carter, Chowdhury,
  Crick, Cunningham, Davies, Davison, Dé, Dennehy, Duan, Dubey, Dwivedi,
  Edwards, Flavián, Gauld, Grover, Hu, Janssen, Jones, Junglas, Khorana,
  Kraus, Larsen, Latreille, Laumer, Malik, Mardani, Mariani, Mithas, Mogaji,
  Nord, O’Connor, Okumus, Pagani, Pandey, Papagiannidis, Pappas, Pathak,
  Pries-Heje, Raman, Rana, Rehm, Ribeiro-Navarrete, Richter, Rowe, Sarker,
  Stahl, Tiwari, {van der Aalst}, Venkatesh, Viglia, Wade, Walton, Wirtz, and
  Wright}]{DWIVEDI2023102642}
\bibinfo{author}{Y.~K. Dwivedi}, \bibinfo{author}{N.~Kshetri},
  \bibinfo{author}{L.~Hughes}, \bibinfo{author}{E.~L. Slade},
  \bibinfo{author}{A.~Jeyaraj}, \bibinfo{author}{A.~K. Kar},
  \bibinfo{author}{A.~M. Baabdullah}, \bibinfo{author}{A.~Koohang},
  \bibinfo{author}{V.~Raghavan}, \bibinfo{author}{M.~Ahuja},
  \bibinfo{author}{H.~Albanna}, \bibinfo{author}{M.~A. Albashrawi},
  \bibinfo{author}{A.~S. Al-Busaidi}, \bibinfo{author}{J.~Balakrishnan},
  \bibinfo{author}{Y.~Barlette}, \bibinfo{author}{S.~Basu},
  \bibinfo{author}{I.~Bose}, \bibinfo{author}{L.~Brooks},
  \bibinfo{author}{D.~Buhalis}, \bibinfo{author}{L.~Carter},
  \bibinfo{author}{S.~Chowdhury}, \bibinfo{author}{T.~Crick},
  \bibinfo{author}{S.~W. Cunningham}, \bibinfo{author}{G.~H. Davies},
  \bibinfo{author}{R.~M. Davison}, \bibinfo{author}{R.~Dé},
  \bibinfo{author}{D.~Dennehy}, \bibinfo{author}{Y.~Duan},
  \bibinfo{author}{R.~Dubey}, \bibinfo{author}{R.~Dwivedi},
  \bibinfo{author}{J.~S. Edwards}, \bibinfo{author}{C.~Flavián},
  \bibinfo{author}{R.~Gauld}, \bibinfo{author}{V.~Grover},
  \bibinfo{author}{M.-C. Hu}, \bibinfo{author}{M.~Janssen},
  \bibinfo{author}{P.~Jones}, \bibinfo{author}{I.~Junglas},
  \bibinfo{author}{S.~Khorana}, \bibinfo{author}{S.~Kraus},
  \bibinfo{author}{K.~R. Larsen}, \bibinfo{author}{P.~Latreille},
  \bibinfo{author}{S.~Laumer}, \bibinfo{author}{F.~T. Malik},
  \bibinfo{author}{A.~Mardani}, \bibinfo{author}{M.~Mariani},
  \bibinfo{author}{S.~Mithas}, \bibinfo{author}{E.~Mogaji},
  \bibinfo{author}{J.~H. Nord}, \bibinfo{author}{S.~O’Connor},
  \bibinfo{author}{F.~Okumus}, \bibinfo{author}{M.~Pagani},
  \bibinfo{author}{N.~Pandey}, \bibinfo{author}{S.~Papagiannidis},
  \bibinfo{author}{I.~O. Pappas}, \bibinfo{author}{N.~Pathak},
  \bibinfo{author}{J.~Pries-Heje}, \bibinfo{author}{R.~Raman},
  \bibinfo{author}{N.~P. Rana}, \bibinfo{author}{S.-V. Rehm},
  \bibinfo{author}{S.~Ribeiro-Navarrete}, \bibinfo{author}{A.~Richter},
  \bibinfo{author}{F.~Rowe}, \bibinfo{author}{S.~Sarker},
  \bibinfo{author}{B.~C. Stahl}, \bibinfo{author}{M.~K. Tiwari},
  \bibinfo{author}{W.~{van der Aalst}}, \bibinfo{author}{V.~Venkatesh},
  \bibinfo{author}{G.~Viglia}, \bibinfo{author}{M.~Wade},
  \bibinfo{author}{P.~Walton}, \bibinfo{author}{J.~Wirtz},
  \bibinfo{author}{R.~Wright},
\newblock \bibinfo{title}{“so what if chatgpt wrote it?” multidisciplinary
  perspectives on opportunities, challenges and implications of generative
  conversational ai for research, practice and policy},
\newblock \bibinfo{journal}{International Journal of Information Management}
  \bibinfo{volume}{71} (\bibinfo{year}{2023}) \bibinfo{pages}{102642}.
  \URLprefix
  \url{https://www.sciencedirect.com/science/article/pii/S0268401223000233}.
  \DOIprefix\doi{https://doi.org/10.1016/j.ijinfomgt.2023.102642}.
\bibitem[{Carr(2023)}]{stackOverflow23}
\bibinfo{author}{D.~F. Carr},
\newblock \bibinfo{title}{{Stack Overflow} is {ChatGPT} casualty: Traffic down
  14\% in {M}arch},
\newblock \bibinfo{journal}{similarweb}  (\bibinfo{year}{2023}). \URLprefix
  \url{https://www.similarweb.com/blog/insights/ai-news/stack-overflow-chatgpt/},
  \bibinfo{note}{last accessed on 21-05-2023}.
\bibitem[{Osman et~al.(2020)Osman, Sierra, Chenu{-}Abente, Shen, and
  Giunchiglia}]{OsmanSCSG20}
\bibinfo{author}{N.~Osman}, \bibinfo{author}{C.~Sierra},
  \bibinfo{author}{R.~Chenu{-}Abente}, \bibinfo{author}{Q.~Shen},
  \bibinfo{author}{F.~Giunchiglia},
\newblock \bibinfo{title}{Open social systems},
\newblock in: \bibinfo{editor}{N.~Bassiliades},
  \bibinfo{editor}{G.~Chalkiadakis}, \bibinfo{editor}{D.~de~Jonge} (Eds.),
  \bibinfo{booktitle}{Multi-Agent Systems and Agreement Technologies - 17th
  European Conference, {EUMAS} 2020, and 7th International Conference, {AT}
  2020, Thessaloniki, Greece, September 14-15, 2020, Revised Selected Papers},
  volume \bibinfo{volume}{12520} of \textit{\bibinfo{series}{Lecture Notes in
  Computer Science}}, \bibinfo{publisher}{Springer}, \bibinfo{year}{2020}, pp.
  \bibinfo{pages}{132--142}. \URLprefix
  \url{https://doi.org/10.1007/978-3-030-66412-1\_9}.
  \DOIprefix\doi{10.1007/978-3-030-66412-1\_9}.
\bibitem[{Osman et~al.(2021)Osman, Chenu{-}Abente, Shen, Sierra, and
  Giunchiglia}]{OsmanCSSG21}
\bibinfo{author}{N.~Osman}, \bibinfo{author}{R.~Chenu{-}Abente},
  \bibinfo{author}{Q.~Shen}, \bibinfo{author}{C.~Sierra},
  \bibinfo{author}{F.~Giunchiglia},
\newblock \bibinfo{title}{Empowering users in online open communities},
\newblock \bibinfo{journal}{{SN} Comput. Sci.} \bibinfo{volume}{2}
  (\bibinfo{year}{2021}) \bibinfo{pages}{338}. \URLprefix
  \url{https://doi.org/10.1007/s42979-021-00714-5}.
  \DOIprefix\doi{10.1007/s42979-021-00714-5}.
\bibitem[{Verhagen et~al.(2018)Verhagen, Neumann, and Singh}]{normsG}
\bibinfo{author}{H.~Verhagen}, \bibinfo{author}{M.~Neumann},
  \bibinfo{author}{M.~P. Singh},
\newblock \bibinfo{title}{Normative multiagent systems: Foundations and
  history},
\newblock in: \bibinfo{editor}{A.~Chopra}, \bibinfo{editor}{L.~van~der Torre},
  \bibinfo{editor}{H.~Verhagen}, \bibinfo{editor}{S.~Villata} (Eds.),
  \bibinfo{booktitle}{Handbook of Normative Multiagent Systems},
  \bibinfo{publisher}{College Publications}, \bibinfo{year}{2018}, p.
  \bibinfo{pages}{3–25}.
\bibitem[{Andrighetto et~al.(2013)Andrighetto, Governatori, Noriega, and
  van~der Torre}]{norms2}
\bibinfo{author}{G.~Andrighetto}, \bibinfo{author}{G.~Governatori},
  \bibinfo{author}{P.~Noriega}, \bibinfo{author}{L.~W.~N. van~der Torre},
  \bibinfo{title}{Normative Multi-Agent Systems}, volume~\bibinfo{volume}{4} of
  \textit{\bibinfo{series}{Dagstuhl Follow-Ups}}, \bibinfo{publisher}{Schloss
  Dagstuhl--Leibniz-Zentrum fuer Informatik}, \bibinfo{address}{Dagstuhl,
  Germany}, \bibinfo{year}{2013}. \URLprefix
  \url{https://drops.dagstuhl.de/opus/volltexte/2013/3997/}.
  \DOIprefix\doi{10.4230/DFU.Vol4.12111}.
\bibitem[{D'Inverno et~al.(2012)D'Inverno, Luck, Noriega, Rodriguez-Aguilar,
  and Sierra}]{2012-D'Inverno}
\bibinfo{author}{M.~D'Inverno}, \bibinfo{author}{M.~Luck},
  \bibinfo{author}{P.~Noriega}, \bibinfo{author}{J.~A. Rodriguez-Aguilar},
  \bibinfo{author}{C.~Sierra},
\newblock \bibinfo{title}{Communicating open systems},
\newblock \bibinfo{journal}{Artificial Intelligence} \bibinfo{volume}{186}
  (\bibinfo{year}{2012}) \bibinfo{pages}{38--94}.
\bibitem[{Georgara et~al.(2020)Georgara, Sierra, and
  Rodr{\'{\i}}guez{-}Aguilar}]{abs-2005-09331}
\bibinfo{author}{A.~Georgara}, \bibinfo{author}{C.~Sierra},
  \bibinfo{author}{J.~A. Rodr{\'{\i}}guez{-}Aguilar},
\newblock \bibinfo{title}{{TAIP:} an anytime algorithm for allocating student
  teams to internship programs},
\newblock \bibinfo{journal}{CoRR} \bibinfo{volume}{abs/2005.09331}
  (\bibinfo{year}{2020}). \URLprefix \url{https://arxiv.org/abs/2005.09331}.
  \href{http://arxiv.org/abs/2005.09331}{{\tt arXiv:2005.09331}}.
\bibitem[{Georgara et~al.(2021)Georgara, Rodr{\'{\i}}guez{-}Aguilar, and
  Sierra}]{GeorgaraRS21}
\bibinfo{author}{A.~Georgara}, \bibinfo{author}{J.~A.
  Rodr{\'{\i}}guez{-}Aguilar}, \bibinfo{author}{C.~Sierra},
\newblock \bibinfo{title}{Towards a competence-based approach to allocate teams
  to tasks},
\newblock in: \bibinfo{editor}{F.~Dignum}, \bibinfo{editor}{A.~Lomuscio},
  \bibinfo{editor}{U.~Endriss}, \bibinfo{editor}{A.~Now{\'{e}}} (Eds.),
  \bibinfo{booktitle}{{AAMAS} '21: 20th International Conference on Autonomous
  Agents and Multiagent Systems, Virtual Event, United Kingdom, May 3-7, 2021},
  \bibinfo{publisher}{{ACM}}, \bibinfo{year}{2021}, pp.
  \bibinfo{pages}{1504--1506}. \URLprefix
  \url{https://www.ifaamas.org/Proceedings/aamas2021/pdfs/p1504.pdf}.
  \DOIprefix\doi{10.5555/3463952.3464140}.
\bibitem[{Georgara et~al.(2022)Georgara, Rodr{\'{\i}}guez{-}Aguilar, Sierra,
  Mich, Kazhamiakin, Aprosio, and Pazzaglia}]{GeorgaraRSMKAP22}
\bibinfo{author}{A.~Georgara}, \bibinfo{author}{J.~A.
  Rodr{\'{\i}}guez{-}Aguilar}, \bibinfo{author}{C.~Sierra},
  \bibinfo{author}{O.~Mich}, \bibinfo{author}{R.~Kazhamiakin},
  \bibinfo{author}{A.~P. Aprosio}, \bibinfo{author}{J.~R. Pazzaglia},
\newblock \bibinfo{title}{An anytime heuristic algorithm for allocating many
  teams to many tasks},
\newblock in: \bibinfo{editor}{P.~Faliszewski}, \bibinfo{editor}{V.~Mascardi},
  \bibinfo{editor}{C.~Pelachaud}, \bibinfo{editor}{M.~E. Taylor} (Eds.),
  \bibinfo{booktitle}{21st International Conference on Autonomous Agents and
  Multiagent Systems, {AAMAS} 2022, Auckland, New Zealand, May 9-13, 2022},
  \bibinfo{publisher}{International Foundation for Autonomous Agents and
  Multiagent Systems {(IFAAMAS)}}, \bibinfo{year}{2022}, pp.
  \bibinfo{pages}{1598--1600}. \URLprefix
  \url{https://www.ifaamas.org/Proceedings/aamas2022/pdfs/p1598.pdf}.
  \DOIprefix\doi{10.5555/3535850.3536047}.
\bibitem[{dos Santos et~al.(2021)dos Santos, Osman, and Schorlemmer}]{thiago1}
\bibinfo{author}{T.~F. dos Santos}, \bibinfo{author}{N.~Osman},
  \bibinfo{author}{M.~Schorlemmer},
\newblock \bibinfo{title}{Learning for detecting norm violation in online
  communities},
\newblock in: \bibinfo{editor}{A.~Theodorou}, \bibinfo{editor}{J.~C. Nieves},
  \bibinfo{editor}{M.~D. Vos} (Eds.), \bibinfo{booktitle}{Coordination,
  Organizations, Institutions, Norms, and Ethics for Governance of Multi-Agent
  Systems {XIV} - International Workshop, {COINE} 2021, London, UK, May 3,
  2021, Revised Selected Papers}, volume \bibinfo{volume}{13239} of
  \textit{\bibinfo{series}{Lecture Notes in Computer Science}},
  \bibinfo{publisher}{Springer}, \bibinfo{year}{2021}, pp.
  \bibinfo{pages}{127--142}. \URLprefix
  \url{https://doi.org/10.1007/978-3-031-16617-4\_9}.
  \DOIprefix\doi{10.1007/978-3-031-16617-4\_9}.
\bibitem[{dos Santos et~al.(2022)dos Santos, Osman, and Schorlemmer}]{thiago2}
\bibinfo{author}{T.~F. dos Santos}, \bibinfo{author}{N.~Osman},
  \bibinfo{author}{M.~Schorlemmer},
\newblock \bibinfo{title}{Ensemble and incremental learning for norm violation
  detection},
\newblock in: \bibinfo{editor}{P.~Faliszewski}, \bibinfo{editor}{V.~Mascardi},
  \bibinfo{editor}{C.~Pelachaud}, \bibinfo{editor}{M.~E. Taylor} (Eds.),
  \bibinfo{booktitle}{21st International Conference on Autonomous Agents and
  Multiagent Systems, {AAMAS} 2022, Auckland, New Zealand, May 9-13, 2022},
  \bibinfo{publisher}{International Foundation for Autonomous Agents and
  Multiagent Systems {(IFAAMAS)}}, \bibinfo{year}{2022}, pp.
  \bibinfo{pages}{427--435}. \URLprefix
  \url{https://www.ifaamas.org/Proceedings/aamas2022/pdfs/p427.pdf}.
  \DOIprefix\doi{10.5555/3535850.3535899}.
\bibitem[{dos Santos et~al.(2023)dos Santos, Cranefield, Savarimuthu, Osman,
  and Schorlemmer}]{thiago3}
\bibinfo{author}{T.~F. dos Santos}, \bibinfo{author}{S.~Cranefield},
  \bibinfo{author}{B.~T.~R. Savarimuthu}, \bibinfo{author}{N.~Osman},
  \bibinfo{author}{M.~Schorlemmer},
\newblock \bibinfo{title}{Cross-community adapter learning (cal) to understand
  the evolving meanings of norm violation},
\newblock in: \bibinfo{booktitle}{32nd International Joint Conference on
  Artificial Intelligence, {IJCAI} 2023, Macao, China, August 19-25, 2023},
  \bibinfo{year}{2023}.

\end{thebibliography}

\end{document}